# Min-Max Similarity: A Contrastive Semi-Supervised Deep Learning Network for Surgical Tools Segmentation

Ange Lou, Kareem Tawfik, Xing Yao, Ziteng Liu, and Jack Noble, *Member, IEEE*

*Abstract*—A common problem with segmentation of medical images using neural networks is the difficulty to obtain a significant number of pixel-level annotated data for training. To address this issue, we proposed a semi-supervised segmentation network based on contrastive learning. In contrast to the previous state-of-the-art, we introduce Min-Max Similarity (MMS), a contrastive learning form of dual-view training by employing classifiers and projectors to build all-negative, and positive and negative feature pairs, respectively, to formulate the learning as solving a MMS problem. The all-negative pairs are used to supervise the networks learning from different views and to capture general features, and the consistency of unlabeled predictions is measured by pixel-wise contrastive loss between positive and negative pairs. To quantitatively and qualitatively evaluate our proposed method, we test it on four public endoscopy surgical tool segmentation datasets and one cochlear implant surgery dataset, which we manually annotated. Results indicate that our proposed method consistently outperforms state-of-the-art semi-supervised and fully supervised segmentation algorithms. And our semi-supervised segmentation algorithm can successfully recognize unknown surgical tools and provide good predictions. Also, our MMS approach could achieve inference speeds of about 40 frames per second (fps) and is suitable to deal with the real-time video segmentation.

## I. INTRODUCTION

Semantic tool segmentation in surgical microscopic or endoscopic video is important for surgical scene understanding for a variety of computer-assisted, robotic-assisted, and minimally invasive surgery (MIS) applications, e.g., [1,44,45]. However, the contents of surgical videos are complex, and robustness of video analysis algorithms is highly affected by different illumination conditions, bleeding, smoke, and occlusions. Recently, deep neural networks have been demonstrated to be the most effective approach for many semantic segmentation tasks. Deep neural networks often leverage a large amount of annotated data for training in order to achieve robust segmentation performance. However, annotations require expert knowledge and thus larger datasets are very expensive and time-consuming to be collected.

Various deep learning architectures have been developed including novel encoder-decoder-based networks [3,4,5], pre-trained feature extractors-based networks [6], and attention-based networks [7,8]. Most of the existing algorithms perform well in a fully-supervised training setting when sizeable annotated data are available. However, in the medical field the scale of available annotated datasets is often limited. To solve the limitation of needing extensive annotated data, some purely unsupervised learning approaches [9,10] have been developed. Generally, the performance of those unsupervised learning approaches is not currently adequate for most medical image segmentation tasks due to their low accuracy. As such, semi-supervised learning (SSL) has become a more promising method. With SSL a limited number of pixel-level annotated images are used for training along with the large quantities of available unlabeled images.

Some recently proposed successful SSL segmentation techniques include mean-teacher-based [14,29], deep co-training [19], cross-pseudo supervision [20], and Duo-SegNet [15]. In the mean teacher method, the model contains a student and a teacher network. During the training, the student network is supervised by two losses: 1). Segmentation loss between labeled predictions and their ground truth; 2). Consistency loss between unlabeled predictions with and without adding noise. The teacher uses the exponential moving average (EMA) weights of the student model. The performance of the mean teacher is highly dependent on the learning ability of the student network. Deep co-training is inspired by the co-training framework to train multiple deep neural networks to be the different views and exploits adversarial examples to encourage view differences. Cross pseudo supervision (CPS) is a consistency regularization approach that is used to encourage high similarity between the predictions of two perturbed networks for the same input image and expand training data by using the unlabeled data with pseudo labels. The Duo-SegNet is also based on multi-view learning [16], the labeled and unlabeled predictions from the segmentation network are fed to critics to analyze the distribution of input

This study was supported in part by grant R01DC014037 from the National Institute for Deafness and other Communication Disorders. This content is solely the responsibility of the authors and does not necessarily represent the official views of this institute.

Ange Lou is with the Electrical and Computer Engineering Department, Vanderbilt University, Nashville, TN 37235 USA (e-mail: ange.lou@vanderbilt.edu).

Kareem Tawfik is with the Otolaryngology-Head and Neck Surgery Department, Vanderbilt University Medical Center, Nashville, TN 37232 USA (e-mail: kareem.tawfik@vumc.org)

Xing Yao is with the Computer Science Department, Vanderbilt University, Nashville, TN 37235 USA (e-mail: xing.yao@vanderbilt.edu).

Ziteng Liu is with the Computer Science Department, Vanderbilt University, Nashville, TN 37235 USA (e-mail: ziteng.liu@vanderbilt.edu).

Jack Noble is with the Electrical and Computer Engineering Department, Vanderbilt University, Nashville, TN 37235 USA (e-mail: jack.noble@vanderbilt.edu).

images. The system is updated by minimizing the sum of the segmentation loss and the adversarial loss.

Contrastive learning plays a dominant role in several computer vision tasks where annotated data are limited. The intuition of this approach is that an input image can be subjected to different transformations but still result in similar feature representations extracted by the neural network encoder and that these feature representations should be dissimilar from those of a different image [11]. A suitable contrastive loss [12,13] is formulated to capture this intuition, and a neural network is training with unlabeled data to minimize this loss. Through contrastive learning, a large amount of unlabeled data can contribute to training.

Recently, numerous contrastive learning based SSL segmentation algorithms [29,30,31] have been proposed for 3D medical image datasets like ACDC [32], Prostate [33] and MMWHS [34,35] to segment organs and lesion areas. However, we find there are fewer works that focus on 2D medical images, especially 2D surgical images. However, numerous works [54] aimed at other 2D image modalities have used small amounts of labelled data to achieve performance on par with fully-supervised methods that use large datasets, e.g., MoCo [25] and SimCLR [26]. The architecture of MoCo is similar to mean teacher, but only one of the networks is updated from the training. SimCLR applies different augmentations for the same images from a mini-batch to build two different views and learn general features. Although these two state-of-the-art methods achieve comparable performance with a fully supervised ResNet-50 [27] on ImageNet, both of them use ResNet-50 with a 12 ResNet blocks, which results in networks that have 375 million parameters. However, for our MMS approach, the two segmentation networks contain 112 million parameters in total.

In this paper, we propose a novel contrastive learning-based semi-supervised technique – Min-Max Similarity (MMS) – to segment the surgical tools. Unlike the existing contrastive learning state of the art methods, which build different views [16] by applying augmentation to original images, we propose a novel concept where all-negative-pairs are used by separating the labeled datasets into two non-overlapping subsets ($X_1 \cap X_2 = \emptyset$), and then the network predicts the representation for an unlabeled image by using the segmentation networks which are trained from these two subsets.

We evaluate our method on four publicly available surgical video datasets, however, our target application is segmentation of the electrode insertion tools in cochlear implant (CI) surgery. CI is considered the standard-of-care treatment for profound sensory-based hearing loss [36,37,38]. With over 700,000 recipients worldwide, CIs are arguably the most successful neural prostheses to date. While speech recognition outcomes with CIs are, on average, remarkable (55% word recognition for unilateral recipients and 65% for bilateral recipients [46,47,48,49,50]), outcomes are highly variable across recipients. For many recipients, speech recognition is elusive and only basic sound awareness is achieved. One factor that has been shown to account for some of the variability in outcomes is the intra-cochlear positioning of the CI electrode array [51,36]. However, in traditional surgery the intra-cochlear electrode position is unknown to the surgeon as the array is threaded into a small opening of the cochlea, and visualization of its final position is obscured by opaque bone. Recent studies aimed at CT-based pre-operative planning of the insertion procedure have shown that improved intra-cochlear placement of the CI can be achieved if the surgeon inserts the array on a pre-operatively planned insertion trajectory [52,53]. Therefore, in this work we aim to use the MMS method we propose to train a neural network for segmentation of the electrode insertion tool in CI surgical video. This represents a critical step towards an image-guidance system that can confirm the surgeon achieves the optimal insertion trajectory for the CI array and provides feedback to the surgeon if trajectory adjustment is needed.

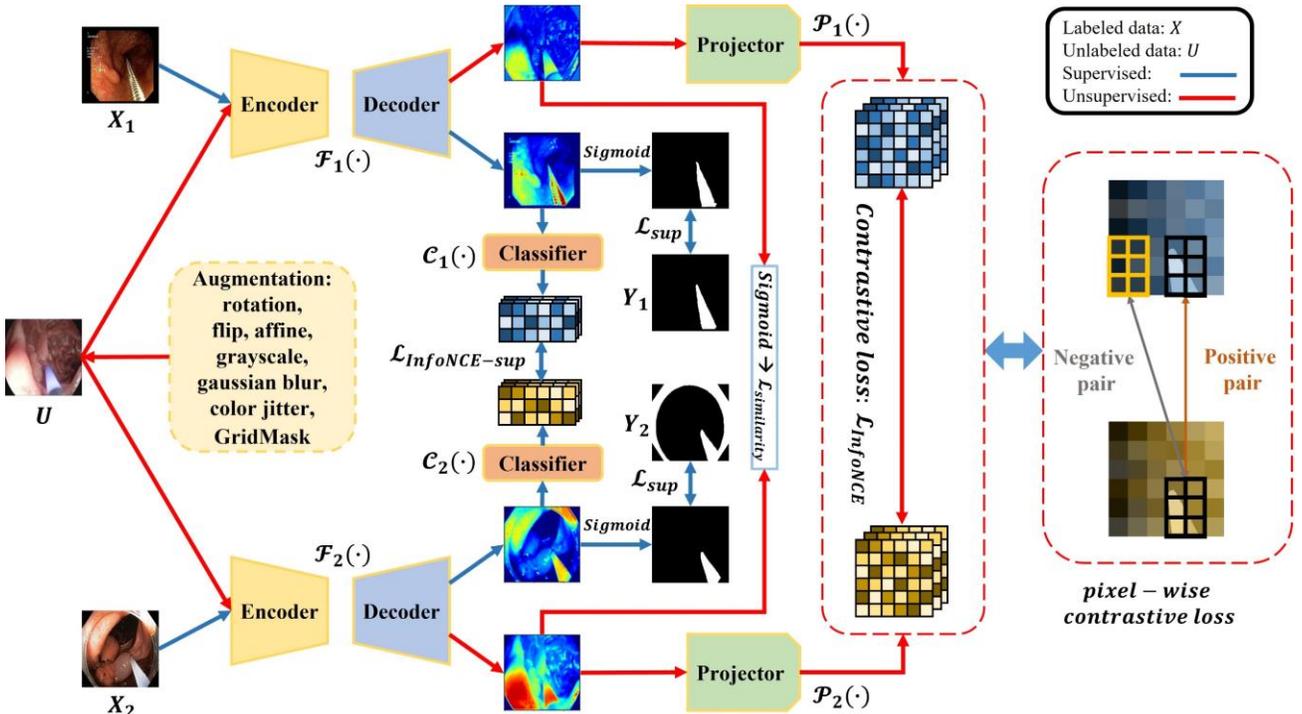

Fig. 1. The architecture of Min-Max Similarity. $\{\mathcal{F}_1, \mathcal{F}_2\}$, $\{\mathcal{C}_1, \mathcal{C}_2\}$ and $\{\mathcal{P}_1, \mathcal{P}_2\}$ denote segmentation networks, classifiers and projectors, respectively. Here, classifiers extract features to build all-negative pairs, and projectors project unlabeled predictions to high dimension features for pixel-wise consistency measurement.

The contributions of our work are highlighted as follows:
- We are the first to use contrastive-learning-based semi-supervised technique to solve the problem of segmenting CI electrode insertion tools from surgery microscope videos.
- We build a CI insertion tool segmentation dataset from 5 different 30 frames-per-second (fps) surgical videos with 40 frames each. This is the first dataset of its kind, to the best of our knowledge.
- We proposed a novel contrastive-learning-based semi-supervised segmentation framework that outperforms the most current state-of-the-art methods on four public surgical tool segmentation datasets and our cochlear implant segmentation dataset.

## II. METHOD

Here, we introduce the **M**in-**M**ax **S**imilarity (MMS) algorithm by providing an overview of the network architecture as shown in Fig. 1. The labeled and unlabeled data are first sent to two segmentation networks. After that, their predicted labels go through a classifier, for labeled data, or a projector, for unlabeled data, for further contrastive analysis. We defined the labeled set as $\mathcal{X} = \{(X_i, Y_i)\}_{i=1}^{m}$, and each pair $(X_i, Y_i)$ consisting of an image $X_i \in \mathbb{R}^{C \times H \times W}$ and its ground truth $Y_i \in \{0,1\}^{H \times W}$. The unlabeled images are assigned as $\mathcal{U} = \{U_i\}_{i=1}^{n}$, where $U_i \in \mathbb{R}^{C \times H \times W}$. These are a set of $n$ unlabeled images where $n \gg m$ ($n$ and $m$ represent number of unlabeled and labeled data). The labeled set is evenly split into two subsets $\{X_i\}_{i=1}^{m} = X_1 \cup X_2$ and $X_1 \cap X_2 = \emptyset$ to ensure the two segmentation networks learn from different views [16]. Segmentation networks, classifiers, and projectors are defined as $\{\mathcal{F}_1, \mathcal{F}_2\}$, $\{\mathcal{C}_1, \mathcal{C}_2\}$, and $\{\mathcal{P}_1, \mathcal{P}_2\}$, respectively.

In our algorithm, we benefit from labeled and unlabeled data. In training, each unlabeled image $U_i$ is sent through both segmentation network and projector paths. Constructing spatial positive and negative pairs from feature representations of the predicted segmentations of the unlabeled images via the projectors permits analyzing the pixel-level consistency in the feature representation. We wish to *minimize* this difference in the feature representation of each unlabeled image across the two segmentation networks and projectors at corresponding spatial positions and to *maximize* the same difference in feature representation for spatial locations that do not correspond. This provides a SSL loss for the unlabeled data. With the labeled data, the two labeled subsets, $X_1$ and $X_2$, are each used in distinct network and classifier paths. While no frames in X1 are present in X2 and vice versa, it is indeed important to clarify that we sample them from the same videos, and it is not a requirement that X1 and X2 be sampled from unique videos. We selected frames for network 1 from the first half of each video and for network 2 from the second half to ensure a substantial time interval exists between the two groups. The classifiers are used to build negative pairs to encourage the networks to learn from different views of labeled data. We wish to *maximize* the difference in classifier features between different labeled images. Our successful training results when considering these to be negative-pairs indicate our separation strategy is reasonable. It is in fact advantageous that contrastive learning can be achieved using negative pairs from the same videos because if this were not the case, more labelled training videos may be required. Due to this dual-objective minimization and maximization optimization design, we refer to our method as Min-Max Similarity (MMS).

**Architecture:** The segmentation networks $\{\mathcal{F}_1, \mathcal{F}_2\}$ share the same encoder-decoder architecture. We leverage Res2Net [17] as an encoder which is pre-trained on ImageNet [18]. The decoder contains four stages, and each stage contains a convolution block and an up-sampling layer, resulting in a predicted segmentation mask at the original image resolution. The filter numbers of the convolution blocks are {512, 256, 64, 1}. The basic element block of the projectors and classifiers are a combination of one convolution layer and one max pooling layer. Here, we choose 3 and 2 blocks for the projectors and classifiers, respectively. The filter numbers for them are {8, 16, 32} and {8, 16}, respectively.

**Loss function:** The total loss function is composed of two pairs of loss terms, one pair of terms is for the labeled data and another for the unlabeled data. For the labeled data, we first compare the prediction with its ground truth by supervised loss $\mathcal{L}_{sup}$. The supervised loss is defined as:

$$\mathcal{L}_{sup} = \mathcal{L}_{IoU}^{w} + \mathcal{L}_{BCE}^{w} \quad (1)$$

where $\mathcal{L}_{IoU}^{w}$ and $\mathcal{L}_{BCE}^{w}$ represent the weighted IoU loss and the weighted binary cross-entropy (BCE) loss. Due to the size and blurred boundary of surgical tools, weights are selected such that pixels with more uncertain classification tend to have higher weight than easy to classify pixels using the methods proposed by Qin et al. and Wei et al. [55,56]. The predicted labels $\{\mathcal{F}_1(X_1), \mathcal{F}_2(X_2)\}$ from the segmentation networks $\{\mathcal{F}_1, \mathcal{F}_2\}$ are sent to classifiers to extract representation features $\{\mathcal{C}_1(\{\mathcal{F}_1(X_1)), \mathcal{C}_2(\mathcal{F}_2(X_2))\}$. Then, the dissimilarity between the two representation features is measured by using sample-wise contrastive loss $\mathcal{L}_{InfoNCE-sup}$, which is defined as:

$$\begin{aligned} L_{InfoNCE-sup} &= -\log \frac{\exp(q^c \cdot k_+^c / \tau)}{\sum_{i=0}^{K} \exp(q^c \cdot k_i^c / \tau)} \\ &= -\log \frac{\exp(0/\tau)}{\sum_{i=0}^{K} \exp(q^c \cdot k_i^c / \tau)} \\ &= -\log \frac{1}{\sum_{i=0}^{K} \exp(q^c \cdot k_i^c / \tau)} \quad (2) \end{aligned}$$

where $q$ and $k$ are feature maps from two classifiers; $\{q^c \cdot k_+^c\}$ is a positive sample pair; $\{q^c \cdot k_i^c\}$ is a negative sample pair; and $\tau$ is the temperature constant we choose to be 0.07 (Note: $X_1 \cap X_2 = \emptyset$ means there are all-negative pairs, thus $q^c \cdot k_+^c = 0$).

For unlabeled data, we firstly send them to segmentation networks with heavy augmentation, and then measure the similarity between two unlabeled predictions $\{\mathcal{F}_1(U), \mathcal{F}_2(U)\}$ using a similarity loss, which is defined as:

$$\mathcal{L}_{similarity} = \mathcal{L}_{sup} = \mathcal{L}_{IoU}^{w} + \mathcal{L}_{BCE}^{w} \quad (3)$$

Then the unlabeled predictions $\{\mathcal{F}_1(U), \mathcal{F}_2(U)\}$ go through their projectors to generate the high-level feature maps $\{\mathcal{P}_1(\mathcal{F}_1(U)), \mathcal{P}_2(\mathcal{F}_2(U))\}$. We use pixel-wise contrastive loss to measure the similarity of the two high-level feature maps which are 1/4 of the original image resolution. The elements at the same spatial location are considered to be positive pairs and others are negative pair. The pixel-wise contrastive loss is measured by:

$$loss_{InfoNCE} = -\log \frac{\exp(q^p \cdot k_+^p / \tau)}{\sum_{i=0}^{K} \exp(q^p \cdot k_i^p / \tau)} \quad (4)$$

where $q$ and $k$ are feature maps from two projectors; $\{q^p \cdot k_+^p\}$ is a positive sample pair; $\{q^p \cdot k_i^p\}$ is a negative sample pair; and $\tau$ is the temperature constant we choose to be 0.07.
Finally, the two segmentation networks are updated simultaneously by the sum of these four losses:

$$L_{total} = \lambda_1 L_{sup} + \lambda_2 L_{InfoNCE-sup}$$
$$+ \lambda_3 L_{similarity} + \lambda_4 L_{InfoNCE} \quad (5)$$

The overall MMS optimization process is summarized in Algorithm 1.

**Data Augmentation:** Contrastive learning generally needs a large amount of augmentation [35]. To enhance the networks' ability to learn general representation features, we apply heavy data augmentation, including rotations, flips, affine transformations, random grayscale noise, gaussian blur, color jitter, and GridMask [24]. For each epoch, every unlabeled image in training set is augmented by those augmentation techniques using Torchvision.

**Implementation Details:** The Min-Max Similarity model is developed in PyTorch [21]. For training the segmentation networks, we use Adam optimizer ($LR = 1e-4$, $\beta_1 = 0.9$ and $\beta_2 = 0.999$).

**Datasets:** We use four public surgical tool segmentation datasets and one cochlear implant dataset developed for this work: 1). Kvasir-instrument [22] consists of 590 annotated frames containing gastrointestinal (GI) procedure tools such as snares, balloons, biopsy forceps, etc. The training set contains 472 images, and the testing set contains 118 images. 2). EndoVis'17 [23] contains 8 robotic surgical videos, each with 255 frames. We select the eighth video as testing set and the other 7 as training sets. It contains more than one type of surgical tool, as shown in Table I. 3). ART-NET [58] consists of 816 annotated laparoscopy images from 29 laparoscopic videos. The training set contains 662 images, and the testing set contains 154 images. 4). RoboTool [59] consists of 514 images extracted from the videos of 20 freely available robotic surgical procedures and annotated for binary segmentation. We randomly select 412 images as training set and the remaining 102 images as testing set. 5). Our cochlear implant (CI) dataset contains 5 surgical videos recorded at 30 frames-per-second (fps) from the surgical microscope. Our training set consists of 183 manually labeled images and 7497 unlabeled images. The testing set contains 40 images from 5 surgical videos with manually labeled ground truth. The ground truth labels were reviewed by an expert clinician from Vanderbilt University Medical Center and confirmed to be accurate. For

TABLE I
DETAILS OF SURGICAL TOOL TYPES IN EndoVis'17.

| Video Number | Types of Surgical Tools |
| --- | --- |
| # 1 | Forceps |
| # 2 | Forceps |
| # 3 | Needle drivers |
| # 4 | Needle drivers and forceps |
| # 5 | Forceps, retractors and vessel sealer |
| # 6 | Needle driver, scissors and forceps |
| # 7 | Forceps and vessel sealer |
| # 8 (test set) | Forceps, retractor, scissors and retractors |

---

**Algorithm 1:** Min-Max Similarity (training)

---

**Input:** Define segmentation networks $\{\mathcal{F}_i(\cdot)\}_{i=1}^{2}$, classifiers $\{\mathcal{C}_i(\cdot)\}_{i=1}^{2}$, projectors $\{\mathcal{P}_i(\cdot)\}_{i=1}^{2}$, batch size $\mathcal{B}$, maximum epoch $E_{max}$, labeled images $\mathcal{X} = \{(X_i, Y_i)\}_{i=1}^{m}$, unlabeled images $\mathcal{U} = \{U_i\}_{i=1}^{n}$, and two labeled sets $\{\mathcal{X}^1, \mathcal{X}^2\}$ where $\mathcal{X}^1 \cup \mathcal{X}^2 \subset \mathcal{X}$ and $\mathcal{X}^1 \cap \mathcal{X}^2 = \emptyset$;

**Output:** parameters $\{\theta_i\}_{i=1}^{2}$ of $\{\mathcal{F}_i(\cdot)\}_{i=1}^{2}$;

**Initialization:** Initialize network, classifier and projector parameters $\{\theta_i\}_{i=1}^{2}$, $\{\mu_i\}_{i=1}^{2}$ and $\{\nu_i\}_{i=1}^{2}$;

**for** $epoch = 1, \dots, E_{max}$ **do**

    **for** $each\ batch\ \mathcal{B}$ **do**

        Generate predictions for labeled data $\mathcal{F}_1(x)$ for all $X_i \in \mathcal{X}^1$, $\mathcal{F}_2(x)$ for all $X_i \in \mathcal{X}^2$ and then for unlabeled data $\mathcal{F}_1(u)$ and $\mathcal{F}_2(u)$ for all $U_i \in \mathcal{U}$;

        Generate classifier and projector feature maps $\mathcal{C}_1(\mathcal{F}_1(x))$, $\mathcal{C}_2(\mathcal{F}_2(x))$, $\mathcal{P}_1(\mathcal{F}_1(u))$, and $\mathcal{P}_2(\mathcal{F}_2(u))$;

        Let $\mathcal{L}_{total} = \lambda_1 \mathcal{L}_{sup} + \lambda_2 \mathcal{L}_{InfoNCE-sup} + \lambda_3 \mathcal{L}_{similarity} + \lambda_4 \mathcal{L}_{InfoNCE}$, as defined in Equations (1) – (4);

        Update $\{\theta_i\}_{i=1}^{2}$ by descending its gradient on $\mathcal{L}_{total}$;

    **end**

**end**

the four public datasets, which are fully labeled, we test our method with 5%, 20%, and 50% of labeled training sets to evaluate the effectiveness of our SSL approach. We resize all images to a resolution of 512 × 288.

**Competing Methods and Evaluation Metrics:** We compare our proposed MMS with the current state-of-the-art methods that include fully supervised UNet, UNet++ [60], TransUNet [42], mean teacher, deep co-training, cross pseudo supervision and Duo SegNet. All approaches are evaluated using the Dice Sørensen coefficient (DSC) [57], mean absolute error (MAE) and F-measure.

**Ablation study:** In our algorithm, we benefit from labeled and unlabeled data through the use of projectors and classifiers. We perform an ablation study to show the effectiveness of including these components in our architecture. Further, we evaluate different combinations of loss weights on the Kvasir-instrument test set. Finally, we divide the augmentation techniques into three groups: 1) random grayscale, colorjitter and gaussian blur; 2) flip, rotation and affine transformations; 3) GridMask. We perform an ablation study to show the improvement from different types of augmentation.

**Generalization to unlabeled tools**: To test the learning abilities of our MMS method for general features of surgical tools that may not be present in the labeled training set, we trained the MMS method with limited portions of the labeled training data of EndoVis'17, where only certain tools were included in the labeled training data. The remainder of the training set that includes other tools was unlabeled. Then we again tested the result on video #8, which includes tools unseen in the labeled training dataset. Specifically, we trained with only the first 5% of the dataset (part of video #1, which includes only forceps) as labeled, then the first 20% (video #1 and part of #2, which includes only forceps) as labeled, and finally the first 50% (videos 1-3 and part of #4, which includes needle drivers in addition to forceps). In all scenarios, the labeled data do not include retractors and scissors present in the testing video.

TABLE II
COMPARISON WITH STATE-OF-THE-ART METHODS ON KVASIR AND ENDOVIS'17.

| Dataset | Method | DSC | | | IOU | | | MAE | | | F-measure | | |
|---|---|---|---|---|---|---|---|---|---|---|---|---|---|
| | UNet (fully) | 0.901 | | | 0.756 | | | 0.027 | | | 0.862 | | |
| | UNet ++ (fully) | 0.893 | | | 0.778 | | | 0.023 | | | 0.875 | | |
| | TransUNet (fully) | **0.905** | | | **0.855** | | | **0.015** | | | **0.922** | | |
| | Label ratio $l_a$ | 5% | 20% | 50% | 5% | 20% | 50% | 5% | 20% | 50% | 5% | 20% | 50% |
| | UNet | 0.706 | 0.730 | 0.799 | 0.435 | 0.508 | 0.606 | 0.075 | 0.055 | 0.043 | 0.609 | 0.674 | 0.755 |
| Kvasir-instrument | UNet++ | 0.567 | 0.736 | 0.823 | 0.440 | 0.612 | 0.720 | 0.085 | 0.041 | 0.028 | 0.440 | 0.760 | 0.837 |
| | TransUNet | 0.541 | 0.753 | 0.867 | 0.252 | 0.706 | 0.845 | 0.093 | 0.029 | 0.015 | 0.402 | 0.826 | 0.916 |
| | Mean Teacher | 0.605 | 0.788 | 0.892 | 0.415 | 0.689 | 0.799 | 0.065 | 0.031 | 0.020 | 0.587 | 0.816 | 0.888 |
| | Deep Co-training | 0.489 | 0.764 | 0.866 | 0.292 | 0.632 | 0.735 | 0.084 | 0.045 | 0.027 | 0.452 | 0.759 | 0.840 |
| | Cross Pseudo | 0.709 | 0.824 | 0.894 | 0.607 | 0.643 | 0.804 | 0.051 | 0.037 | 0.020 | 0.755 | 0.783 | 0.891 |
| | Duo-SegNet | 0.403 | 0.834 | 0.861 | 0.274 | 0.701 | 0.755 | 0.081 | 0.033 | 0.026 | 0.430 | 0.824 | 0.860 |
| | **Min-Max Similarity (ours)** | **0.776** | **0.874** | **0.925** | **0.650** | **0.768** | **0.873** | **0.043** | **0.024** | **0.013** | **0.787** | **0.868** | **0.932** |
| | UNet (fully) | 0.894 | | | 0.840 | | | 0.027 | | | 0.912 | | |
| | UNet ++ (fully) | **0.909** | | | **0.841** | | | 0.026 | | | **0.914** | | |
| | TransUNet (fully) | 0.904 | | | 0.826 | | | 0.029 | | | 0.905 | | |
| | Label ratio $l_a$ | 5% | 20% | 50% | 5% | 20% | 50% | 5% | 20% | 50% | 5% | 20% | 50% |
| | UNet | 0.823 | 0.869 | 0.885 | 0.653 | 0.772 | 0.819 | 0.057 | 0.040 | 0.029 | 0.784 | 0.872 | 0.902 |
| EndoVis' 17 | UNet++ | 0.825 | 0.882 | 0.890 | 0.651 | 0.743 | 0.760 | 0.058 | 0.044 | 0.041 | 0.788 | 0.853 | 0.864 |
| | TransUNet | 0.837 | 0.873 | 0.882 | 0.713 | 0.775 | 0.790 | 0.047 | 0.039 | 0.035 | 0.833 | 0.875 | 0.882 |
| | Mean Teacher | 0.875 | 0.901 | 0.910 | 0.797 | 0.848 | 0.849 | 0.037 | 0.028 | 0.024 | 0.885 | 0.915 | 0.920 |
| | Deep Co-training | 0.848 | 0.895 | 0.895 | 0.777 | 0.845 | 0.847 | 0.038 | 0.026 | 0.026 | 0.875 | 0.913 | 0.917 |
| | Cross Pseudo | 0.886 | 0.909 | 0.913 | 0.813 | 0.850 | 0.855 | 0.029 | 0.025 | 0.021 | 0.895 | 0.919 | 0.926 |
| | Duo-SegNet | 0.879 | 0.906 | 0.912 | 0.806 | 0.849 | 0.864 | 0.033 | 0.025 | 0.023 | 0.893 | 0.918 | 0.927 |
| | **Min-Max Similarity (ours)** | **0.909** | **0.931** | **0.940** | **0.861** | **0.890** | **0.899** | **0.023** | **0.018** | **0.017** | **0.925** | **0.942** | **0.947** |
| | UNet (fully) | 0.894 | | | 0.752 | | | 0.029 | | | 0.859 | | |
| | UNet ++ (fully) | **0.908** | | | 0.799 | | | 0.023 | | | 0.888 | | |
| | TransUNet (fully) | 0.904 | | | **0.823** | | | **0.019** | | | **0.903** | | |
| | Label ratio $l_a$ | 5% | 20% | 50% | 5% | 20% | 50% | 5% | 20% | 50% | 5% | 20% | 50% |
| | UNet | 0.660 | 0.713 | 0.812 | 0.521 | 0.510 | 0.679 | 0.072 | 0.062 | 0.038 | 0.685 | 0.676 | 0.809 |
| ART-NET | UNet++ | 0.717 | 0.761 | 0.866 | 0.590 | 0.600 | 0.743 | 0.053 | 0.051 | 0.030 | 0.742 | 0.750 | 0.852 |
| | TransUNet | 0.685 | 0.764 | 0.841 | 0.628 | 0.633 | 0.733 | 0.047 | 0.043 | 0.032 | 0.773 | 0.775 | 0.846 |
| | Mean Teacher | 0.747 | 0.835 | 0.889 | 0.614 | 0.726 | 0.814 | 0.051 | 0.033 | 0.021 | 0.761 | 0.841 | 0.897 |
| | Deep Co-training | 0.726 | 0.820 | 0.875 | 0.629 | 0.714 | 0.811 | 0.049 | 0.033 | 0.021 | 0.772 | 0.833 | 0.895 |
| | Cross Pseudo | 0.759 | 0.824 | 0.874 | 0.629 | 0.708 | 0.797 | 0.047 | 0.035 | 0.023 | 0.772 | 0.829 | 0.887 |
| | Duo-SegNet | 0.738 | 0.771 | 0.833 | 0.608 | 0.664 | 0.729 | 0.048 | 0.039 | 0.032 | 0.756 | 0.798 | 0.843 |
| | **Min-Max Similarity (ours)** | **0.784** | **0.869** | **0.917** | **0.652** | **0.758** | **0.843** | **0.045** | **0.029** | **0.017** | **0.790** | **0.863** | **0.915** |
| | UNet (fully) | 0.786 | | | 0.617 | | | 0.088 | | | 0.763 | | |
| | UNet ++ (fully) | 0.807 | | | 0.656 | | | 0.068 | | | 0.792 | | |
| | TransUNet (fully) | **0.808** | | | **0.672** | | | **0.063** | | | **0.804** | | |
| | Label ratio $l_a$ | 5% | 20% | 50% | 5% | 20% | 50% | 5% | 20% | 50% | 5% | 20% | 50% |
| | UNet | 0.516 | 0.661 | 0.730 | 0.413 | 0.505 | 0.558 | 0.133 | 0.105 | 0.093 | 0.584 | 0.671 | 0.716 |
| RoboTool | UNet++ | 0.500 | 0.691 | 0.734 | 0.397 | 0.527 | 0.568 | 0.152 | 0.098 | 0.087 | 0.561 | 0.690 | 0.724 |
| | TransUNet | 0.516 | 0.718 | 0.732 | 0.497 | 0.549 | 0.559 | 0.123 | 0.087 | 0.090 | 0.664 | 0.709 | 0.717 |
| | Mean Teacher | 0.575 | 0.742 | 0.784 | 0.443 | 0.637 | 0.679 | 0.137 | 0.074 | 0.061 | 0.614 | 0.773 | 0.809 |
| | Deep Co-training | 0.519 | 0.714 | 0.752 | 0.397 | 0.593 | 0.636 | 0.143 | 0.080 | 0.068 | 0.568 | 0.744 | 0.777 |
| | Cross Pseudo | 0.559 | 0.711 | 0.758 | 0.429 | 0.593 | 0.641 | 0.147 | 0.083 | 0.069 | 0.601 | 0.745 | 0.781 |
| | Duo-SegNet | 0.586 | 0.701 | 0.746 | 0.488 | 0.556 | 0.647 | 0.117 | 0.086 | 0.070 | 0.656 | 0.715 | 0.786 |
| | **Min-Max Similarity (ours)** | **0.646** | **0.781** | **0.831** | **0.544** | **0.697** | **0.750** | **0.104** | **0.058** | **0.046** | **0.705** | **0.821** | **0.857** |

**Cochlear implant dataset cross validation:** Our goal is to design a reliable method to segment the cochlear implant insertion tools in unseen surgical videos. Thus, we also perform a leave-one-out cross-validation (LOOCV), where we leave out one video as testing set and use the four other videos as a training set. We also compared the segmentation performance of our proposed method with other four semi-supervised segmentation algorithms with the same LOOCV approach.

## III. RESULTS

The qualitative and quantitative results comparison of the proposed method to four state-of-the-art methods are shown in Fig. 2, Table II, and Table III. The results reveal that the proposed method (MMS) outperformed the other selected state-of-the-art methods on the five datasets. This is especially true for the small-scale Kvasir-instrument dataset, which only contains 472 labeled training samples. The two segmentation networks of MMS were only supervised by 12 labeled images each (5% of Kvasir) but achieved 7%-37% improvement

TABLE III
COMPARISON WITH STATE-OF-THE-ART METHODS ON COCHLEAR IMPLANT.

| Dataset | Method | DSC | IOU | MAE | F-measure |
|---|---|---|---|---|---|
| Cochlear | UNet (fully) | 0.863 | 0.812 | 0.043 | 0.896 |
| | UNet ++ (fully) | **0.888** | **0.840** | 0.033 | **0.913** |
| | TransUNet (fully) | 0.882 | 0.835 | **0.031** | 0.910 |
| | Label ratio $l_a$ | 2.4% (183/7497) | | | |
| | Mean Teacher | 0.914 | 0.854 | 0.032 | 0.920 |
| | Deep Co-training | 0.847 | 0.825 | 0.048 | 0.902 |
| | Cross Pseudo | 0.910 | 0.850 | 0.028 | 0.919 |
| | Duo-SegNet | 0.869 | 0.834 | 0.035 | 0.910 |
| | **Min-Max Similarity (ours)** | **0.920** | **0.861** | **0.021** | **0.925** |

TABLE IV
ABLATION STUDY
(a). NETWORK ARCHITECTURE

| Label ratio | Classifier | Projector | DSC | IOU | MAE | F-measure |
|---|---|---|---|---|---|---|
| 5 % | ✗ | ✗ | 0.641 | 0.476 | 0.056 | 0.645 |
| | ✓ | ✗ | 0.713 | 0.550 | 0.051 | 0.710 |
| | ✗ | ✓ | 0.720 | 0.553 | 0.051 | 0.713 |
| | ✓ | ✓ | **0.776** | **0.650** | **0.043** | **0.787** |
| 20% | ✗ | ✗ | 0.720 | 0.617 | 0.040 | 0.763 |
| | ✓ | ✗ | 0.787 | 0.649 | 0.038 | 0.787 |
| | ✗ | ✓ | 0.823 | 0.659 | 0.035 | 0.794 |
| | ✓ | ✓ | **0.874** | **0.768** | **0.024** | **0.868** |
| 50% | ✗ | ✗ | 0.814 | 0.712 | 0.029 | 0.832 |
| | ✓ | ✗ | 0.887 | 0.796 | 0.021 | 0.887 |
| | ✗ | ✓ | 0.896 | 0.804 | 0.020 | 0.891 |
| | ✓ | ✓ | **0.925** | **0.873** | **0.013** | **0.932** |

(b) HYPER-PARAMETER ANALYSIS FOR LOSS WEIGHTS

| $\lambda_1$ | $\lambda_2$ | $\lambda_3$ | $\lambda_4$ | DSC | IOU | MAE | F-measure |
|---|---|---|---|---|---|---|---|
| 0.1 | 0.2 | 0.3 | 0.4 | 0.919 | 0.846 | 0.015 | 0.917 |
| 0.1 | 0.2 | 0.4 | 0.3 | 0.912 | 0.838 | 0.016 | 0.912 |
| 0.2 | 0.1 | 0.3 | 0.4 | 0.920 | 0.850 | 0.015 | 0.919 |
| 0.2 | 0.1 | 0.4 | 0.3 | 0.923 | 0.854 | 0.014 | 0.921 |
| 0.3 | 0.4 | 0.1 | 0.2 | 0.914 | 0.845 | 0.016 | 0.916 |
| 0.3 | 0.4 | 0.2 | 0.1 | 0.916 | 0.841 | 0.016 | 0.914 |
| 0.4 | 0.3 | 0.1 | 0.2 | 0.921 | 0.856 | 0.014 | 0.922 |
| 0.4 | 0.3 | 0.2 | 0.1 | 0.918 | 0.846 | 0.015 | 0.916 |
| 0.2 | 0.2 | 0.3 | 0.3 | 0.917 | 0.836 | 0.016 | 0.911 |
| 0.3 | 0.3 | 0.2 | 0.2 | 0.919 | 0.850 | 0.015 | 0.919 |
| **0.25** | **0.25** | **0.25** | **0.25** | **0.925** | **0.873** | **0.013** | **0.932** |

(C) AUGMENTATION

| Group 1 | Group 2 | Group 3 | DSC | IOU | MAE | F-mesure |
|---|---|---|---|---|---|---|
| ✗ | ✗ | ✗ | 0.916 | 0.846 | 0.015 | 0.917 |
| ✓ | ✗ | ✗ | 0.919 | 0.852 | 0.015 | 0.920 |
| ✓ | ✓ | ✗ | 0.921 | 0.866 | 0.013 | 0.926 |
| ✓ | ✓ | ✓ | 0.925 | 0.873 | 0.013 | 0.932 |

TABLE V
COMPARISON WITH STATE-OF-ART METHODS ON REORGINAZIED ENDOVIS'17

| Dataset | Method | DSC | | | IOU | | | | MAE | | | F-measure |
|---|---|---|---|---|---|---|---|---|---|---|---|---|
| | UNet (fully) | 0.894 | | | 0.840 | | | | 0.027 | | | 0.912 |
| | UNet ++ (fully) | **0.909** | | | **0.841** | | | | **0.026** | | | **0.914** |
| | TransUNet (fully) | 0.904 | | | 0.826 | | | | 0.029 | | | 0.905 |
| EndoVis' 17 | Label ratio $l_a$ | 5% | 20% | 50% | 5% | 5% | 20% | 50% | 5% | 5% | 20% | 50% | 5% |
| | UNet | 0.761 | 0.710 | 0.822 | 0.624 | 0.761 | 0.710 | 0.822 | 0.624 | 0.761 | 0.710 | 0.822 | 0.624 |
| | UNet++ | 0.754 | 0.732 | 0.827 | 0.594 | 0.754 | 0.732 | 0.827 | 0.594 | 0.754 | 0.732 | 0.827 | 0.594 |
| | TransUNet | 0.754 | 0.762 | 0.868 | 0.627 | 0.754 | 0.762 | 0.868 | 0.627 | 0.754 | 0.762 | 0.868 | 0.627 |
| | Mean Teacher | 0.782 | 0.790 | 0.878 | 0.665 | 0.782 | 0.790 | 0.878 | 0.665 | 0.782 | 0.790 | 0.878 | 0.665 |
| | Deep Co-training | 0.737 | 0.734 | 0.843 | 0.637 | 0.737 | 0.734 | 0.843 | 0.637 | 0.737 | 0.734 | 0.843 | 0.637 |
| | Cross Pseudo | 0.768 | 0.777 | 0.832 | 0.654 | 0.768 | 0.777 | 0.832 | 0.654 | 0.768 | 0.777 | 0.832 | 0.654 |
| | Duo-SegNet | 0.814 | 0.788 | 0.879 | 0.712 | 0.814 | 0.788 | 0.879 | 0.712 | 0.814 | 0.788 | 0.879 | 0.712 |
| | **Min-Max Similarity (ours)** | **0.837** | **0.838** | **0.921** | **0.733** | **0.837** | **0.838** | **0.921** | **0.733** | **0.837** | **0.838** | **0.921** | **0.733** |

TABLE VI
COMPARISON WITH STATE-OF-ART METHODS ON LOOCV VALIDATION

| Dataset | Method | DSC | | | | | IOU | | | | |
|---|---|---|---|---|---|---|---|---|---|---|---|
| | | Fold 1 | Fold 2 | Fold 3 | Fold 4 | Fold 5 | Fold 1 | Fold 2 | Fold 3 | Fold 4 | Fold 5 |
| | UNet (fully) | 0.745 | **0.557** | **0.791** | 0.743 | **0.808** | 0.674 | **0.464** | **0.673** | 0.617 | **0.697** |
| | UNet++ (fully) | **0.747** | 0.520 | 0.771 | **0.746** | 0.783 | **0.679** | 0.412 | 0.663 | **0.632** | 0.675 |
| | TransUNet (fully) | 0.737 | 0.512 | 0.757 | 0.723 | 0.749 | 0.663 | 0.405 | 0.645 | 0.601 | 0.616 |
| | Mean Teacher | 0.794 | 0.600 | 0.709 | 0.745 | 0.752 | 0.688 | 0.457 | 0.576 | 0.642 | 0.629 |
| | Deep Co-training | 0.742 | 0.554 | 0.739 | 0.745 | 0.741 | 0.669 | 0.459 | 0.642 | 0.615 | 0.622 |
| | Cross Pseudo | 0.819 | 0.659 | 0.768 | 0.734 | 0.770 | 0.744 | 0.558 | 0.668 | 0.611 | 0.646 |
| | Duo SegNet | 0.748 | 0.597 | 0.690 | 0.729 | 0.800 | 0.680 | 0.494 | 0.611 | 0.607 | 0.688 |
| | MMS (ours) | **0.874** | **0.728** | **0.843** | **0.807** | **0.875** | **0.784** | **0.579** | **0.743** | **0.675** | **0.783** |
| Cochlear | Method | MAE | | | | | F-measure | | | | |
| | | Fold 1 | Fold 2 | Fold 3 | Fold 4 | Fold 5 | Fold 1 | Fold 2 | Fold 3 | Fold 4 | Fold 5 |
| | UNet (fully) | 0.064 | **0.093** | 0.052 | 0.061 | 0.060 | **0.795** | 0.628 | **0.805** | 0.775 | **0.822** |
| | UNet++ (fully) | **0.058** | 0.104 | 0.061 | 0.062 | **0.058** | 0.792 | 0.583 | 0.793 | 0.768 | 0.797 |
| | TransUNet (fully) | 0.068 | 0.112 | 0.064 | 0.072 | 0.063 | 0.753 | 0.562 | 0.781 | 0.734 | 0.770 |
| | Mean Teacher | 0.053 | 0.088 | 0.075 | 0.064 | 0.062 | 0.815 | 0.628 | 0.731 | 0.782 | 0.772 |
| | Deep Co-training | 0.064 | 0.091 | 0.063 | 0.067 | 0.064 | 0.802 | 0.630 | 0.797 | 0.762 | 0.767 |
| | Cross Pseudo | 0.044 | 0.079 | 0.058 | 0.065 | 0.059 | 0.853 | 0.716 | 0.801 | 0.727 | 0.785 |
| | Duo SegNet | 0.062 | 0.084 | 0.075 | 0.068 | 0.056 | 0.809 | 0.662 | 0.759 | 0.756 | 0.815 |
| | MMS (ours) | **0.037** | **0.072** | **0.046** | **0.053** | **0.039** | **0.879** | **0.733** | **0.853** | **0.806** | **0.878** |

compared with the other four methods. For each increment of included labeled data, MMS resulted in the best segmentation performance. On the Kvasir-instrument dataset with 50% of labeled images, MMS reached 0.925 DSC, which was 2.4% and 3.1% higher than fully supervised UNet and a more recent state-of-the-art method, Cross Pseudo Supervision, respectively. Also, MMS was more robust than other methods as can be visualized from the second column in Fig. 2. MMS was less affected by the bubbles, which makes the prediction contour (blue line) more accurately fit the surgical tools. The advantage of our MMS is much more obvious on the results of EndoVis'17 datasets. We randomly selected 5%, 20% and 50% of labeled data from cases #1 to #7 in the training set to serve as our labeled training set and case #8 as the unlabeled training set. It is clear that the MMS has stronger general-feature-learning ability. MMS only needed 5% labeled training data to exceed fully supervised performance. The other state-of-art methods needed at least 20% of the labeled data to get similar results. The performance of MMS improved with increasing increments of labeled data for training, and these results surpass the performance of fully supervised and other semi-supervised methods. On ART-NET dataset, MMS outperforms the other four state-of-the-art semi-supervised methods on all three label-ratio training sets. In Fig. 2, it can be observed that only MMS could predict a complete and accurate contour for the tool under dark illumination condition. On the RoboTool dataset, MMS achieved 0.831 DSC, which was 4.7%-8.5% improvement compared with other four methods. And in Fig. 2, the contour MMS predicted is more accurate than all fully supervised method especially the slender part of the right surgical tool. On our cochlear implant dataset, there is only one type of tool, the CI insertion tool, so all of the SSL methods can achieve reasonable results. However, only MMS could predict a complete and accurate contour for our tool. From Fig. 2, it can be observed that MMS predicts the most accurate contour for surgical tools regardless of their size, type, and number. Furthermore, we also report the IOU, MAE and F-measure score for all semi and fully supervised method. It is clear that our MMS achieved best performance on all five datasets. Compared with the most recent advanced UNet, MMS has about 1-2% improvement with 50% of labeled images.

The results of ablation study are shown in Table IV. In addition to verifying the contributions of the projectors and classifiers, we evaluated different combinations of loss weights $\lambda_1$, $\lambda_2$, $\lambda_3$ and $\lambda_4$ for experiments and determine the best combination. All experiments in Table IV are conducted for Kvasir-instrument dataset with 50% of annotated data. Finally, we can find when four loss has equal weights, the MMS can achieve best performance.

From the Table IV (a), it can be observed that adding classifiers and projectors help the dual-view architecture learn more general features and achieve the best results with 5%, 20% and 50% labeled training images. With only classifiers, we can make sure the network learning from different views but cannot accurately measure the consistency of predictions for unlabeled data. With only projectors, we have consistent feature representation for the predictions, but the network does not learn general features. When we use both, classifier and projector work together to help the segmentation networks outperforms the other methods. When the labeled data increase from 20 to 50%, the labeled training set includes two types of tools instead of just one, but the labeled data still do not include the retractors and scissors present in the testing set. However, the MMS method has the ability to learn the general features of surgical tools necessary to segment those tools for which ground truth labels were not provided. With 50% of labeled data included, the MMS method is the only one to outperform the fully-supervised segmentation accuracy.

The results of the LOOCV study on the cochlear implant

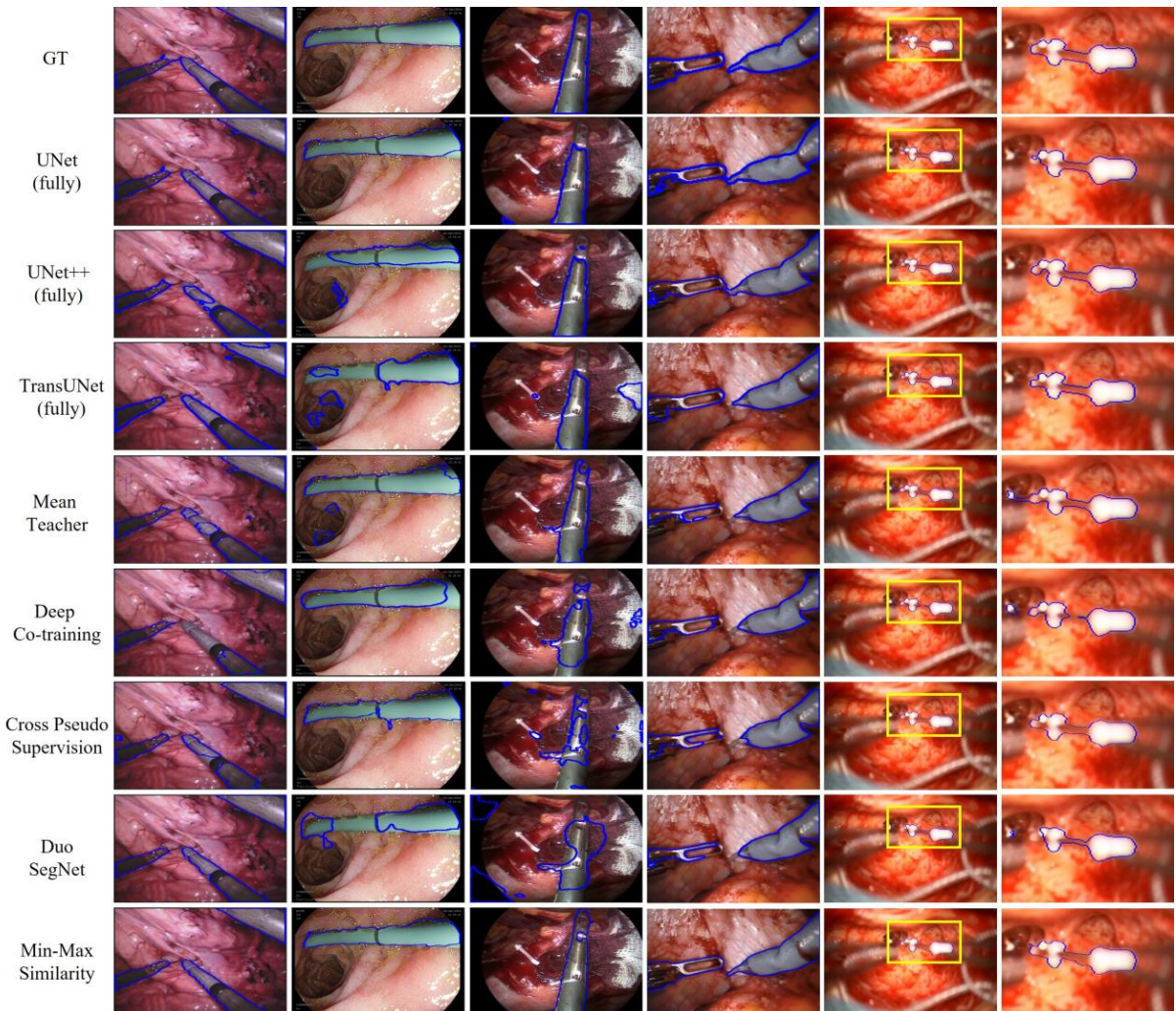

Fig. 2. Visual comparison of our method with state-of-the-art models. Segmentation results are shown for 50% of labeled training data for Kvasir-instrument, EndVis'17, ART-NET and RoboTool, and 2.4% labeled training data for cochlear implant. **From left to right are EndoVis'17, Kvasir-instrument, ART-NET, RoboTool, Cochlear implant and region of interest (ROI) of Cochlear implant.**

learn a more general representation of the labeled and unlabeled data.

From the Table IV (c), it can be observed that the improvement from different augmentation groups. Apply color change (group 1) or physical transformations (group 2) can slightly improve the performance. Using advanced augmentation (group 3) can achieved better improvement.

Table V presents the results of our experiment to test how well the MMS method generalizes to tools that are not present in the labeled dataset. As can be seen in the table, for each proportion of training data included, our MMS method insertion tool segmentation dataset are shown in Table VI. Here fold 1 to 5 represent using each of the videos #1 to #5 as the left-out test case, respectively. From the results, we can see that our MMS outperforms than other state-of-the-art methods. For the most challenge video (#2), the tool is more perpendicular to the camera than in the other videos, which makes the region of interest smaller and the illumination and reflection conditions different than in the other four videos. However, MMS can predict a reasonable segmentation mask for the insertion tool. Overall, MMS masks are about 7%~17% more accurate than the other semi-supervised methods.

**Inference speed:** The speed testing experiment is performed with one RTX A5000 GPU, CUDA 11.3 and cuDNN V8.2.1 on the Pytorch platform. Averaging over 100 frames, which is widely used to measure the real-time semantic segmentation [61, 62], we achieve 40 frames per second (FPS) inference speed with 512 x 288 RGB images.

## IV. DISCUSSION AND CONCLUSIONS

We proposed a contrastive-learning-based algorithm for semi-supervised surgical tools segmentation and demonstrated its effectiveness on publicly available datasets and our own CI dataset. Compared with the existing semi-supervised and contrastive learning methods, which focus on how to measure the consistency between unlabeled predictions, we proposed the all-negative pair concept to encourage the networks to learn from different views and capture real general features. Also, we use pixel-wise contrastive loss to measure the pixel-level consistency in features extracted from the unlabeled data to enhance the segmentation performance. From the experiments, MMS more accurately segments tools that are not included in the labeled training data than other methods. Also, MMS could achieve fully supervised performance by only using small amounts of labeled training data, especially in our cochlear implant dataset.

Unlike the mean teacher, the backbone of our MMS is a dual-view segmentation network, where both segmentation networks are participating in training and share information with each other. Further, in the mean teacher method, only one type of augmentation is used – adding noise. MMS adds more extensive perturbations to unlabeled images to make sure more general features can be learned. Compared with deep co-training and Duo-SegNet, MMS does not focus heavily on analysis of predicted labels. MMS places more focus on building positive and negative feature pairs to compute the consistency between two feature representations of unlabeled predictions. To avoid misleading the networks during the training, we fully use the properties of our labeled subsets ($X_1 \cap X_2 = \emptyset$) and contrastive learning to build the all-negative pairs and to ensure the networks learn from different views. For the CPS method, pseudo labels participate in supervision, which is not suitable for the small-scale medical datasets because errors in the pseudo labels potentially mislead the networks when there are not enough training data.

To the best of our knowledge, there do not exist any methods for automatic segmentation of the CI electrode insertion tool in surgical microscope video. Such methods could be critical for image-guided CI electrode insertion procedures. In this work, we built a segmentation dataset from clinical CI surgical videos to fill this gap. Also, we proposed a robust and accurate semi-supervised segmentation method to solve the limitations of lack of extensive annotated images from the surgical videos and proved its performance on different datasets. When the goal is to process surgical videos in close to real-time, inference speed is an important factor when design networks. Lots of high-accuracy methods like vision transformer (ViT) based networks [41,42,43] are not suitable for real-time video segmentation due to its large scale. However, one benefit of the MMS learning strategy is that the segmentation network uses a simple encoder-decoder based architecture, where inference speed can reach 40 FPS with a 512 × 288 RGB image.

The MMS can still be improved by addressing potential bias between two labeled subsets, e.g., it is possible that one network sees a tool that another does not. Also, we could consider further use of labeled data. For example, the labeled data could be used two ways in training, where in addition to having the network learn from two different views $\{X^1, X^2\}$, it is also possible to send each image down both network paths like our unlabeled data to build positive pairs after going through classifier and to build additional positive-negative pairs after the projectors to enhance learning pixel-level consistency. These limitations and variations will be considered in our future work.